# Title: Mapping small reservoirs across Brazil from 1984 to 2025

**Authors:** Kylen Solvik[1,2*], Luis Gustavo Carvalho[3], Marcia N. Macedo[2,4,5]

**Affiliations:**

[1]Department of Geography, University of Colorado Boulder, Boulder, CO 80309, USA

[2]Department of Ecology, Evolution, and Environmental Biology, Columbia University, New York, NY 10027, USA

[3]State University of Mato Grosso (UNEMAT), Nova Xavantina, MT, Brazil

[4]Woodwell Climate Research Center, Falmouth, MA 02540

[5]Columbia Climate School, Columbia University, New York, NY 10027, USA

*Corresponding author. Email: kylen.solvik@columbia.edu

**Abstract**

Water research in Brazil largely overlooks the widespread damming of small streams for agricultural uses including watering cattle, farm-scale hydropower, irrigation, and aquaculture. These ubiquitous dams and their reservoirs affect water temperature, stream connectivity, aquatic habitats, greenhouse gas emissions, and evaporative water losses. Mapping small reservoirs is challenging because it requires reliably detecting small water bodies and distinguishing artificial reservoirs from natural lakes. As a result, most regional and global datasets exclude them. To address this gap, we trained a deep learning computer vision model to accurately segment small (0.1 to 100 ha), stream-fed, surface water reservoirs in Brazil leveraging data from Landsat 5–9. Applying our model from 1984 to 2025, we created annual reservoir maps for the entire country to evaluate how their count, size, and distribution have changed over time. The number of detected reservoirs grew nearly fourfold from 224,059 to 817,458, while their total surface area increased from 3490 $km^2$ to 8455 $km^2$. To our knowledge, this is the first country-wide annual dataset representing the evolution of small reservoirs over four decades. The publicly available annual maps highlight the extent and cumulative impacts of small stream impoundments across Brazil, providing actionable insights for managing freshwater ecosystems and water resources.

## 1. Introduction

### *1.1 Agricultural reservoirs in Brazil*

Brazil has more freshwater than any other country on Earth, but this abundance can obscure critical water issues, especially relating to agriculture (Dobrovolski and Rattis, 2015; Schulz and Ioris, 2017). While clearing of primary forests for cattle pasture and croplands has attracted intense scrutiny (Barona et al., 2010; Morton et al., 2006; Pendrill et al., 2022), the impacts of agriculture on water have received comparatively little attention, despite growing evidence that climate change and growing water demand pose threats to both water security and agricultural production in Brazil (Ballarin et al., 2023; de Araújo et al., 2004; Hirata et al., 2025; Nóbrega et al., 2011; Perazzoli et al., 2013; Rattis et al., 2021). Addressing this challenge will require reliable data on the state of water resources in Brazil and how they change in response to large-scale agricultural expansion and intensification.

One oft-overlooked impact of land-use change on water systems is the widespread construction of small dams that now dot Brazil's agricultural landscapes. The resulting reservoirs—typically used for providing water for cattle or farm-scale hydropower and, less frequently, for fish aquaculture or crop irrigation (Arvor et al., 2018; Macedo et al., 2013)—have a total estimated surface area of 51,000 ha in the state of Mato Grosso alone (Lathuillière et al., 2019). Despite their small size, these reservoirs can have large cumulative impacts. Studies have demonstrated effects on downstream water temperature and water quality (Chandesris et al., 2019; Fencl et al., 2015; Habets et al., 2014; Macedo et al., 2013; Zaidel et al., 2021); hydrological connectivity (Barbarossa et al., 2020; Belletti et al., 2020; Fencl et al., 2015; Kibler and Tullos, 2013; J. Sun et al., 2024); evaporative water loss (Lathuillière et al., 2019; Tanny et al., 2008); biodiversity (Barbarossa et al., 2020; Lange et al., 2018); and greenhouse gas emissions (Grinham et al., 2018; Odebiri et al., 2024; Ollivier et al., 2019).

Understanding how agriculture impacts water sustainability requires accurately and comprehensively mapping these reservoirs and their evolution over time, yet identifying small stream barriers in satellite imagery remains challenging (Sun et al., 2025). Although several studies have attempted to map small earthen dams over specific regions of Brazil (Arvor et al., 2018; Lathuillière et al., 2019; Macedo et al., 2013; Souza et al., 2019), they have been limited by the difficulty of distinguishing reservoirs from natural lakes and wetlands with remote sensing. As a result, there is no consistent, long-term dataset on small dam distribution in Brazil today, but see ANA (2020) for a compilation of the best available data for Brazil.

To fill this gap, we trained a deep learning semantic segmentation model to develop annual maps of small human-made reservoirs across Brazil from 1984 to 2025 using Landsat 5, 7, 8, and 9 imagery. Our approach provides a consistent, multi-decadal record of small reservoir expansion at a continental scale, enabling analysis of the potential past, present, and future drivers of reservoir construction and use.

### *1.2. Detection and classification of small water bodies via remote sensing*

Accurate mapping of agricultural reservoirs is challenging due to their small size as well as spectral and morphological similarities to natural water bodies. A trained human can identify them based on shape (e.g., an unnaturally straight downstream edge) and contextual clues (e.g., proximity to pasture, cropland, or buildings), but manual annotation at scale is time-intensive and impractical. Agricultural reservoirs in Brazil are frequently smaller than 1 ha, resulting in mixed pixels and vegetation that obscures the surface. Small waterbodies also vary temporally and sometimes disappear entirely, further complicating detections, particularly when there are gaps in the sensor record. Nevertheless, machine learning approaches have shown promise for water mapping tasks (Lathuillière et al., 2019; Pekel et al., 2016; Souza et al., 2019), especially deep learning/artificial intelligence models (Chen et al., 2018; Gharbia, 2023; Jonnala et al., 2025; Luo et al., 2022; Mullen et al., 2023; D. Sun et al., 2024; Wieland et al., 2023; Yuan et al., 2021).

Past approaches to waterbody classification from remotely sensed imagery often involve a two-stage process: (1) mapping all surface water; and (2) classifying waterbodies by type via morphological, spectral, or temporal information. One approach for stage 2 is to apply random forest models to distinguish natural and human-made waterbodies based on size, shape, and surrounding land use/cover (Lathuillière et al., 2019; MapBiomas, 2025; Souza et al., 2019). While it can be effective, this approach propagates errors from the first stage to the second. Another method leverages temporal information, classifying waterbodies that appear abruptly and persist over time as human-made (Arvor et al., 2018; Zhang et al., 2019). There are several key issues. First, it assumes that all new waterbodies appearing in the landscape are artificial, when in fact natural waterbodies are quite dynamic. For instance, playa lakes in the western United States may be dry for years before filling suddenly (Bartuszevige et al., 2012; Solvik et al., 2021). Second, it assumes uniform patterns of dam creation, maintenance, and disappearance that may not hold over large, socially and environmentally heterogeneous areas. Finally, it requires satellite records pre-dating dam construction, which effectively restricts detection to reservoirs built in the last 42 years (since the launch of Landsat 5). In Brazil, reservoir creation far predates the satellite record. For example, the imperial government directed the widespread construction of dams in northeast Brazil after the Grande Seca (“Big Drought”) in 1877-79 (Cavalcante et al., 2022; Greenfield, 1992).

In lieu of the two-stage approaches described above, we trained a Multi-Scale Attention Network (MA-Net) (Fan et al., 2020) to perform binary semantic segmentation, which delineates full reservoir extents. Neural networks have been shown to excel at learning from diverse information to achieve high accuracy on similar tasks (Feng et al., 2019; Gui et al., 2024; Guo et al., 2020; Kim et al., 2019). In addition to overall high accuracy, the MA-Net semantic segmentation approach has three other important advantages. First, it does not rely on strict assumption-based rules to distinguish reservoirs from other water objects. Second, by proceeding in a single analysis step, it avoids the propagation of error inherent in two-stage approaches. Finally, once trained, the lightweight model can be quickly applied over large areas, which we leverage here to create annual reservoir maps covering all of Brazil from 1984 to 2025.

## 2. Methods

### *2.1. Training and evaluation data*

To create reservoir masks for training and evaluation, we annotated 1314 randomly selected 5km x 5km (500 x 500-pixel at 10m resolution) tiles from a 2017 cloud-free Sentinel-2 composite, leveraging Sentinel's higher native resolution (10m) to enhance annotation. Using LabelBox ("Labelbox," 2025) , we manually outlined every reservoir in each tile, labeling 4474 reservoirs in all. To improve model performance in difficult cases (e.g., oxbow lakes and small rivers), we added 196 tiles over river floodplains and wetlands without reservoirs. Since these were not randomly selected, we only used them for training and excluded them from the validation and test sets to avoid biasing our accuracy evaluation.

For training and inference, we created Landsat 5 Enhanced Thematic Mapper (ETM, 1984 – 2012), Landsat 7 ETM+ (1999 – 2024), Landsat 8 Operational Land Imager (OLI, 2013 – 2025), and Landsat 9 OLI-2 (2022 – 2025) cloud-free median-value surface reflectance composites for all of Brazil using Google Earth Engine (Gorelick et al., 2017). We used the six bands that are most consistent between Landsat generations: Blue, Green, Red, NIR, SWIR-1, and SWIR-2. After tile annotation, we extracted the corresponding Landsat 8 tiles at their native 30m resolution as well as resampled to 10m resolution using bicubic and bilinear interpolation to evaluate the effect of interpolation on model performance. After data normalization, we divided the annotated training data into two strata: images without reservoirs ($n$ = 775) and images containing at least one reservoir ($n$ = 539). These strata were then split into training (984 images total, including the 196 "floodplains" images), validation (263), and test (263) sets. Details of mask annotation and data preprocessing are available in the Supplemental Material.

*2.2. Training and evaluation*

We used an MA-Net (Fan et al., 2020) with a ResNet-34 backbone (He et al., 2016) as implemented in Segmentation Models Pytorch (Iakubovskii, 2019), trained from scratch without pretrained weights on the annotated Landsat 8 tiles. To provide more context for the neural network, 640 x 640-pixel images were used as inputs, and the outputs were clipped to 500 x 500 pixels to match the annotated masks. Training data were augmented by applying random vertical and horizontal flips and random 90-degree rotations, all with probability 0.5. We trained the model using an Adam optimizer with a Dice loss function and a learning rate of $1e^{-4}$, an exponential learning rate decay of 0.9 applied after every epoch, weight decay of $2e^{-6}$, and early stopping based on validation-set Intersection-over-Union (IoU) with patience of 5 epochs. Final accuracy was then assessed on the withheld test set of 263 images using pixel-wise and object-based precision, recall, F1, and IoU statistics.

To improve consistency across Landsat generations, for each satellite we tuned and applied an independent binary prediction threshold (pixels with model scores above the threshold were classified as reservoir pixels, pixels below were non-reservoir). For Landsat 8, we chose the threshold that minimized the difference between precision and recall on the validation set (0.704). We calibrated Landsat 7 to Landsat 8 via the same process, but with the Landsat 8 predictions as "ground truth", using 2017 (the year of the annotated data). Finally, we repeated the process to calibrate Landsat 5 to Landsat 7 using the 2010 (the last composite with full Landsat 5 availability) Landsat 7 validation-set predictions as ground truth. For Landsat 9, we used the same threshold as Landsat 8 since the OLI-2 sensor aboard Landsat 9 is an enhanced copy of Landsat 8's OLI. This led to prediction thresholds of 0.704, 0.704, 0.003, and 0.036 for Landsat 9, 8, 7, and 5, respectively. Using these thresholds, we ran model inference over all Brazil for Landsat 5 from 1984 - 2011, Landsat 7 from 2000 - 2024, Landsat 8 from 2014 – 2025, and Landsat 9 from 2022 – 2025 (see Supplemental Material for more details).

## 3. Results

### *3.1. Model accuracy*

The MA-Net model performed well on the challenging task of reservoir segmentation, with high pixel-wise IoU, F1, precision, and recall on the withheld test set of 263 annotated images from 2017 (Table 1; Table S1 contains accuracy statistics on the validation set). The Landsat data upsampled to 10-m resolution via bicubic interpolation (F1-score = 71.3) outperformed both bilinear interpolation (F1-score = 71.0) and predicting at native 30-m resolution (F1 = 68.1), and so we used bicubic interpolation for the final predictions. Object-based precision (78.7) was higher than recall (67.1), suggesting that our final reservoir counts are likely conservative (Table S2). Reservoirs smaller than 0.1 ha (10 pixels at 10m resolution) were removed from the final predictions due to low accuracy (precision = 32.6, recall = 6.9; Table S2). Finally, the IoU scores between sensors (81.7 for Landsat 8/9, 80.2 for Landsat 7/8, and 74.4 for Landsat 5/7) indicate high agreement on images in the withheld test set.

**Table 1.** Accuracy statistics on withheld 2017 test set, comparing sensors and resampling methods.

| | **Landsat 8 (10m, bicubic)** | **Landsat 8 (10m, bilinear)** | **Landsat 8 (native 30m)** | **Landsat 7 (10m, bicubic)** |
|---|---|---|---|---|
| **IoU-Score** | 55.4 | 55.0 | 51.6 | 54.1 |
| **F1-Score** | 71.3 | 71.0 | 68.1 | 70.2 |
| **Precision** | 69.1 | 67.0 | 64.1 | 67.4 |
| **Recall** | 73.6 | 75.5 | 72.6 | 73.3 |

*3.2. Trends in reservoir count and area*

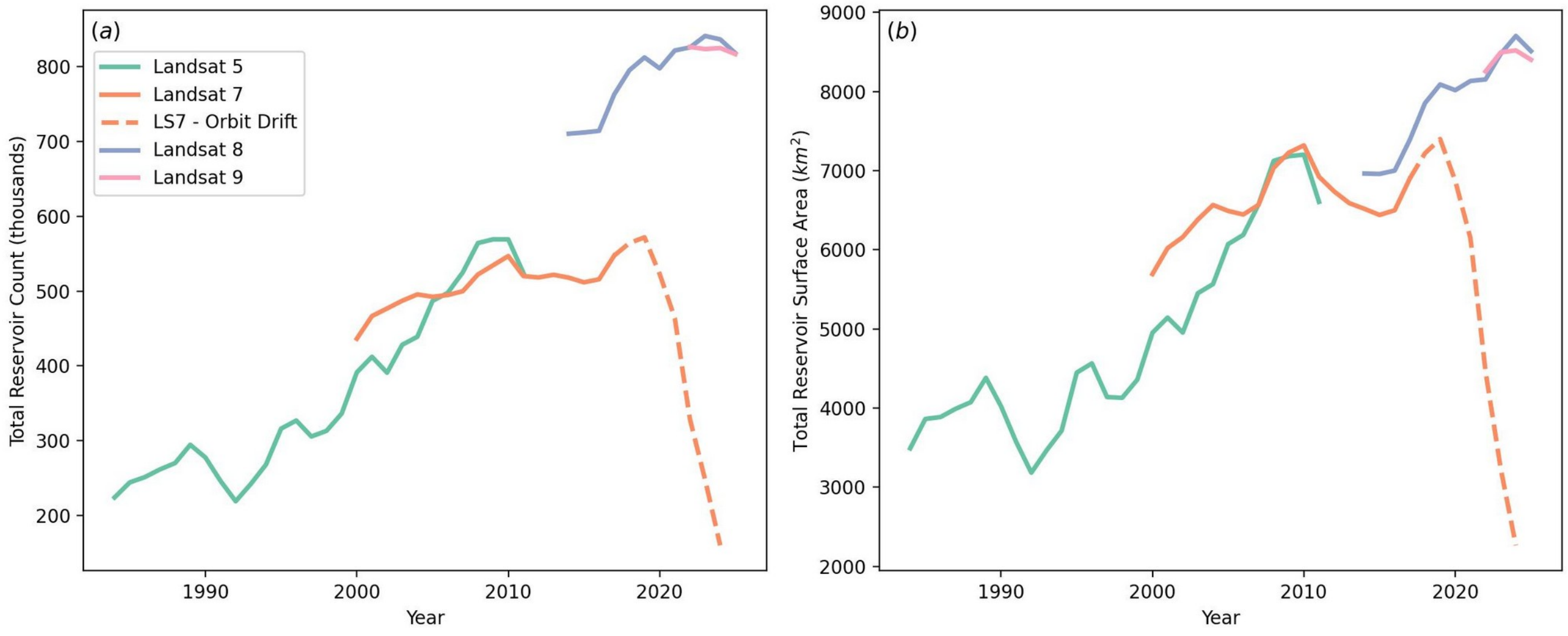

**Figure 1. Evolution of small reservoirs in Brazil from 1984 to 2025.** (a) Reservoir count and (b) total surface area over the Landsat record. Colors denote the different Landsat sensors, indicating periods of overlap. Note that Landsat 7's acquisition times became earlier over time due to orbital drift (dashed orange line), which led to decreasing surface reflectance after 2017 (Qiu et al., 2021).

The count and total surface area of reservoirs more than doubled during the study period (Figure 1; Table S3 contains exact counts). In 1984, the beginning of the Landsat 5 record, there were 224,059 reservoirs with a total surface area of 3490 $km^2$. By 2025, based on Landsat 8, there were 817,458 reservoirs with a total surface area of 8455 km$^2$—approximately equivalent to the surface area of Lake Titicaca, the largest freshwater lake in South America.

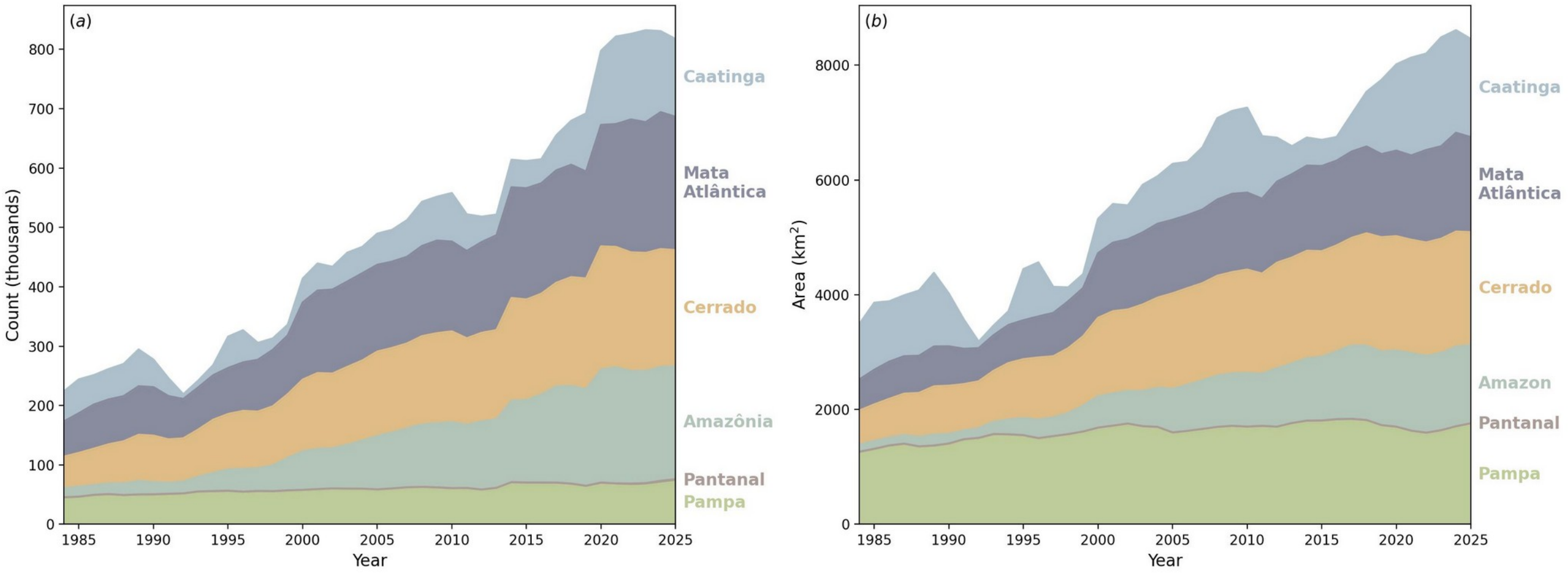

**Figure 2. Reservoir count and area by biome.** In years when two Landsat sensors were operational, the count and area reflect the mean of the sensors (Landsat 5 and 7 from 2000 – 2011; Landsat 7 and 8 from 2014 – 2019; and Landsat 8 and 9 from 2022 – 2025). We excluded Landsat 7 for 2020 and beyond, given its marked orbit drift.

The overall upward trend in reservoir count and area obscures distinct regional patterns, shown in Figures 2 and 3. At the beginning of the study period, reservoirs were concentrated in four of Brazil's six biomes: the Pampa, Cerrado, Mata Atlântica, and Caatinga. The Pampa alone contained almost half of the total reservoir area, while the Amazônia and the Pantanal had very few reservoirs. By the mid-1990s, the distribution began to change. While the Pampa maintained roughly the same number and area of reservoirs throughout the study period, the count and area in Amazônia began to increase rapidly. By 2025, Amazônia had 238 thousand reservoirs with a total area of 1364 km$^2$, a 991% area increase from 1984 (125 km$^2$)—by far the largest growth among the six biomes. Reservoirs also increased in the Cerrado and Mata Atlântica, but the Pantanal remained relatively free of reservoirs.

The semi-arid, drought-prone Caatinga displayed the highest variability over time, suggesting that reservoirs may shrink and even disappear during drought events.

### *3.3. Comparing across Landsat generations*

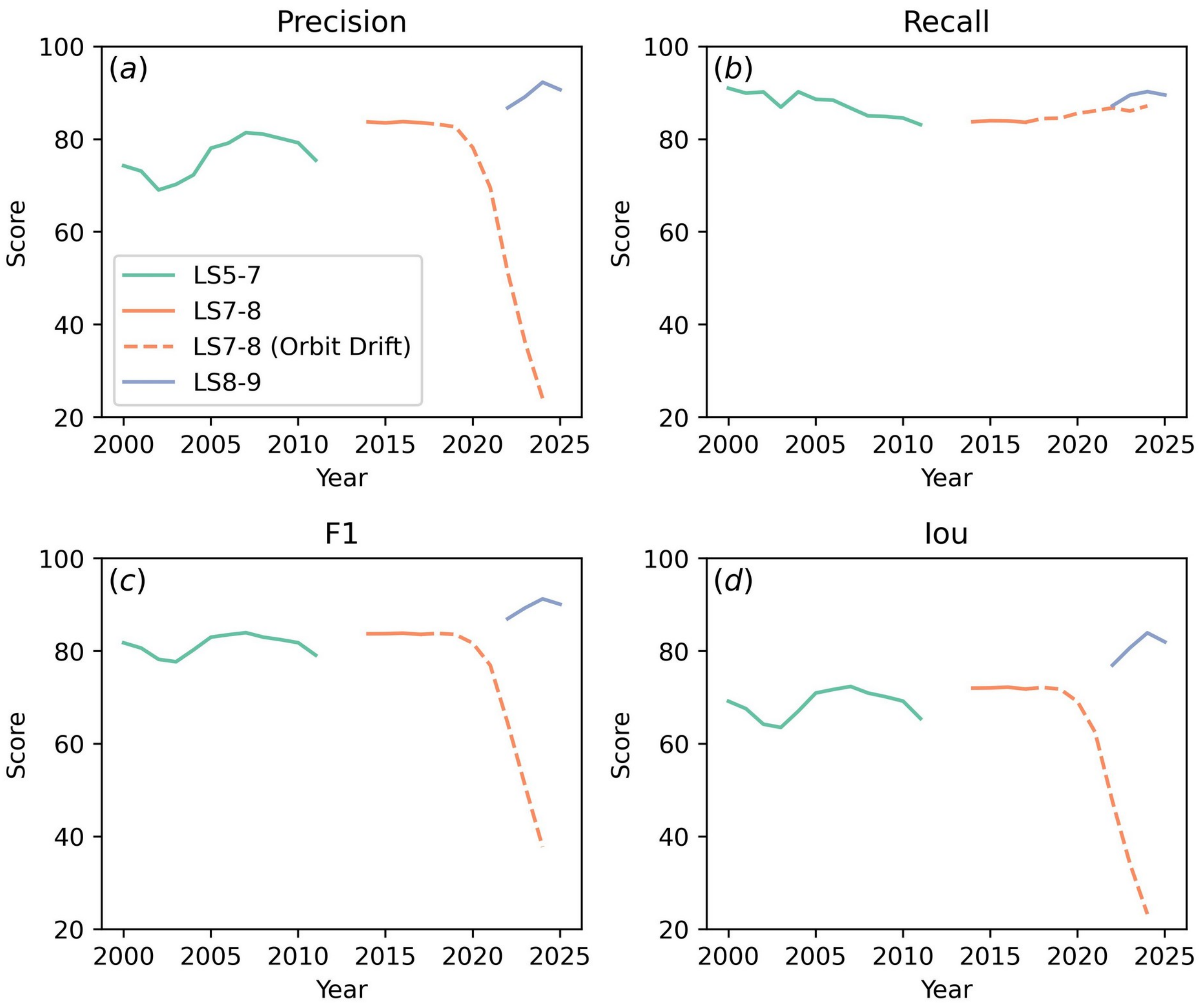

**Figure 4. Accuracy statistics for pairwise sensor comparisons during periods of overlap:** (a) precision, (b) recall, (c) F1-score, and (d) Intersection-over-Union.

Comparing the full predicted maps shows general agreement between the Landsat sensors, but there are some notable differences (Figure 4, Table 2). While Landsat 7 identified just 6.7% more reservoirs by count than Landsat 5, it mapped 14.9% more reservoir surface area on average.

Surprisingly, Landsat 7's Scanline Corrector Failure in 2003 (which resulted in the loss of roughly 20% of each scene) did not affect the agreement between Landsat 5 and Landsat 7 (Figure 4).

**Table 2. Comparison of final maps between Landsat generations in overlapping years.** During periods of overlap, we report the average value of maps from the two sensors. The earlier satellite (Landsat 8 for 2022 – 2025, Landsat 7 for 2014 – 2019, Landsat 5 for 2000 – 2011) was treated as the "ground truth" for the precision and recall statistics.

| **Time Period** | ***2022 - 2025*** | | ***2014 - 2019**** | | ***2000 - 2011*** | |
|---|---|---|---|---|---|---|
| **Satellites** | ***LS9*** | ***LS8*** | ***LS8*** | ***LS7*** | ***LS7*** | ***LS5*** |
| **Reservoir Count (thousands)** | 822.9 | 830.4 | 751.2 | 538.1 | 497.7 | 483.1 |
| **Reservoir Area ($km^2$)** | 8416.7 | 8459.7 | 7375.4 | 6828.9 | 6568.4 | 6083.2 |
| **IoU** | 80.9 | | 72.8 | | 69.8 | |
| **F1** | 89.4 | | 84.2 | | 81.3 | |
| **Precision** | 89.7 | | 81.3 | | 79.3 | |
| **Recall** | 89.1 | | 87.4 | | 85.5 | |

*Landsat 7 data after 2019 was excluded from the comparisons due to effects of its orbit drift.

Landsat 7 and Landsat 8 had even higher agreement (mean IoU = 72.0 and F1 = 83.7%) with more balanced precision (83.4%) and recall (84.1%), but identified 8 detected considerably more reservoirs. On average, Landsat 8 detected 43.8% more reservoirs than Landsat 7 during the same years, but just 1.9% more area (Table 2). We explored two possible explanations: (A) Landsat 8 identified many very small reservoirs that Landsat 7 did not detect, or (B) Landsat 8 identified the same reservoirs as Landsat 7 but fragmented them into multiple smaller segments. For every year with overlapping data, we classified each reservoir as "shared"—meaning it intersected with at least one reservoir in the other satellite's data for the same year—or "unshared". We then averaged the shared and unshared counts and areas over the period from 2014 to 2019. If Landsat 8 systematically identified more very small reservoirs than Landsat 7 (explanation A), we expect the "unshared" count of Landsat 8 to be large and its "unshared" area to be quite small. If instead both sensors identified the same reservoirs, but Landsat 8 fragmented them into multiple segments (explanation B), we expect the "shared" count of Landsat 8 to be higher than Landsat 7 but with similar "unshared" results. Results show that Landsat 8 had many more unique ("unshared") reservoirs, but a fairly small relatively small "unshared" area (Figure 5), suggesting that Landsat 8 identified more very-small reservoirs than Landsat 7, thus increasing its reservoir count while minimally impacting total area.

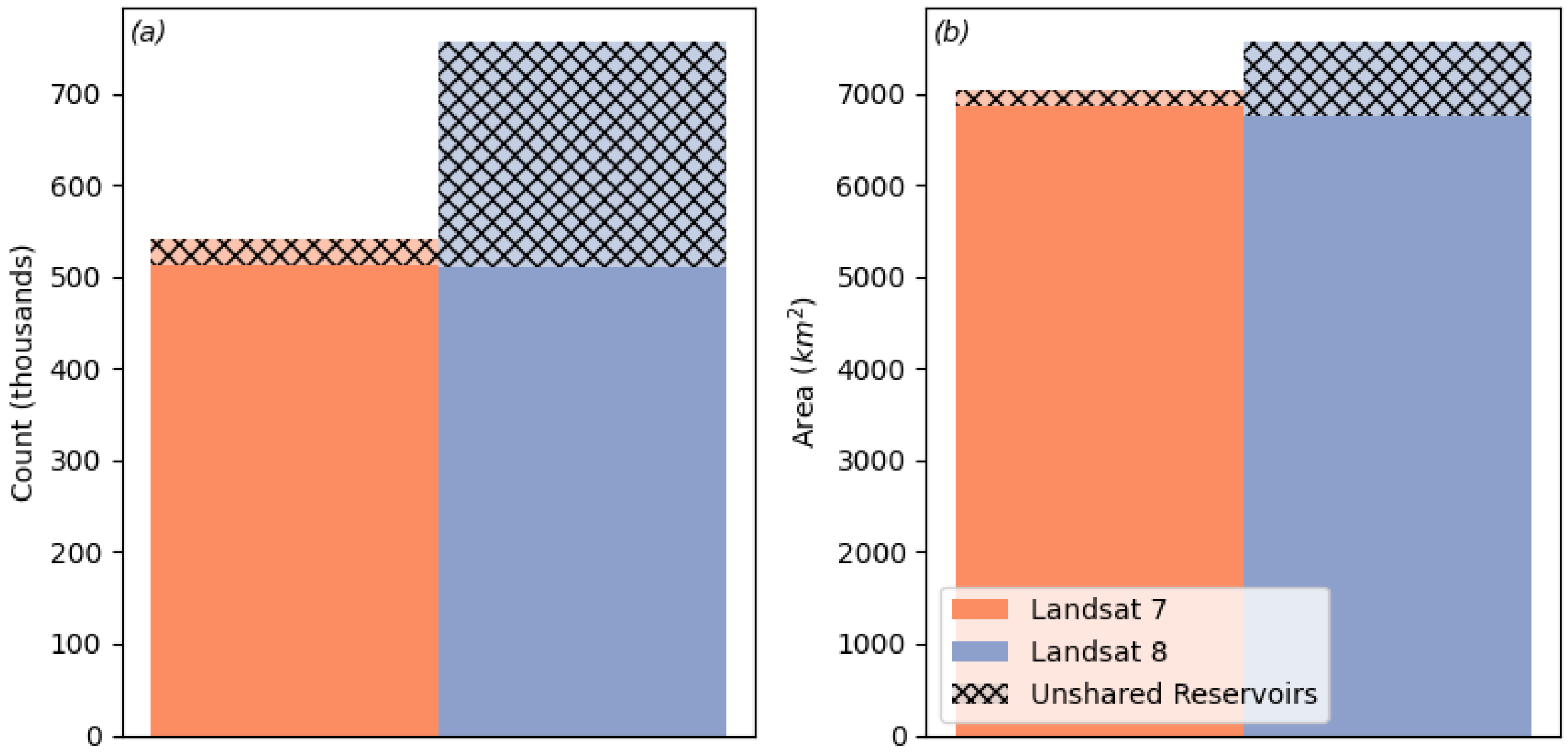

**Figure 5. Contributions of shared and unshared reservoirs between Landsat 7 and 8.** Hatching indicates the proportion of the count (a) and area (b) that was unique (unshared) to each sensor's detections during the overlapping period from 2014-2019.

Landsat 8 and Landsat 9 had the highest agreement in reservoir count and area, as well as IoU (80.8 for the overlap period from 2022 – 2025). This high but not perfect agreement is consistent with previous studies comparing their surface reflectance (Trevisiol et al., 2024; Xu et al., 2024).

## 4. Discussion

### *4.1. Explaining regional trends: drought and agricultural expansion*

Reservoir creation over the past 42 years was closely coupled with agricultural expansion, adding a new dimension to the story of Brazil's journey to become an agricultural superpower—a story that has typically focused on global trade, land clearing, road construction, and colonization programs but rarely considers their hydrological implications (Fearnside et al., 2013; Ioris, 2017; Jepson et al., 2010; Oliveira and Hecht, 2016; Rausch and Gibbs, 2016). The areas with the highest reservoir creation since 1984 (Figure 3) correspond to recent agricultural frontiers, particularly in the Amazon and Cerrado biomes, which saw the largest proportional increase in reservoir count and area (Figure 2). Interestingly, while deforestation has seen substantial increases and decreases over the years (Fearnside, 2005, 1982; Silva Junior et al., 2021), the rate of reservoir creation stayed relatively constant throughout the entire study period, with occasional spikes, plateaus, and dips. The number and area of reservoirs stabilized briefly in the early 2010s—a period with the lowest deforestation rates in decades (F. G. Assis et al., 2019), but rose again after 2016 (Figures 1 and 2). Reservoir counts appear to have decreased slightly in the past two years, but more time is needed to determine if that is the beginning of a trend or simply a temporary dip due to the strong 2023-2024 El Niño or other short-term factors.

Jumps in reservoir counts in 2000, 2014, and 2019 are likely due to changes in satellite data availability (corresponding to the introduction of Landsat 7, the introduction of Landsat 8, and the last year of Landsat 7, respectively), but other discontinuities in the time series appear to align with periods of drought, particularly in the Caatinga biome. This arid region has a history of government-funded programs to build dams for water security dating back to the Grande Seca ("Big Drought") in the 1870s (Cavalcante et al., 2022; Greenfield, 1992) and displays the most variability over time of any of the biomes (Figure 2). The Caatinga receives by far the lowest and most variable rainfall of all Brazilian biomes, and observed dips in reservoir count appear to correspond to years of extreme drought, as seen in 1992, 1998, 2002, 2010, and 2012-2015 (Brito et al., 2018; Marengo et al., 2017). Given the high aridity, Caatinga reservoirs may dry up more often than other regions only to be later refilled after heavy rainfall, creating more intra- and inter-year variation. Overall, the region displayed unique patterns of reservoir creation and disappearance that warrant further investigation, especially given the region's high vulnerability to climate change (Marengo et al., 2017).

### *4.2. Applications and implications for science across Landsat generations*

Our annual reservoirs maps can accelerate vital research on the impacts of small dams on methane and carbon dioxide emissions (DelSontro et al., 2018; Holgerson and Raymond, 2016; Ollivier et al., 2019); hydrological connectivity (Barbarossa et al., 2020; Belletti et al., 2020; Fencl et al., 2015); stream water temperature (Chandesris et al., 2019; Macedo et al., 2013; Zaidel et al., 2021); biodiversity (Barbarossa et al., 2020; Lange et al., 2018); and water balance in agricultural systems (Lathuillière et al., 2025; Mady et al., 2020). For example, recent research has highlighted the outsized greenhouse gas emissions from small water bodies, especially small artificial reservoirs, but these studies have been forced to focus on limited regions where detailed reservoir information was available (Ollivier et al., 2019); lump natural and human-made water bodies together (Holgerson and Raymond, 2016; Pi et al., 2022) despite evidence that gas fluxes are much higher from artificial impoundments (Peacock et al., 2021; Zhang et al., 2026); or approximate reservoir count and surface area using power-law estimates (DelSontro et al., 2018). With location and surface area for each reservoir, our dataset can refine these crucial emissions estimates across Brazil. Temporal dynamics are important, as emissions from lakes and impoundments are expected to rise throughout the 21$^{st}$ century due to increased eutrophication (Beaulieu et al., 2019). Our annual time series enables new insights into patterns of dam building in relation to agricultural production, weather, and policy over the past four decades, and can be updated annually to monitor changes in the face of climate change and rapidly increasing irrigation in Brazil (ANA, 2024).

Depending on the application, users should consider filtering our reservoir maps to specific years and sensors to produce the best results. While area estimates were consistent across all four sensors after tuning the binary prediction thresholds, Landsat 8 and 9 detected more reservoirs than their predecessors (Table 2). For cases where area is the primary concern, such as greenhouse gas flux or evaporative water loss estimation, the full time series from all four satellites can be analyzed together. For applications requiring the highest agreement in reservoir count over time, such as stream fragmentation analyses, the best data depends on the time period: Landsat 5 and 7 provide the greatest continuity from 1984 to 2019, while Landsat 8 and 9 demonstrate superior sensitivity to small reservoirs for analyses from 2014 until today. Due to its drifting orbit, Landsat 7 should not be used after 2019, and these results are excluded from our final datasets.

Sensor improvements likely explain Landsat 8 and 9's increased reservoir detections relative to their predecessors. The OLI sensors' narrower band-widths, particularly in the shortwave- and near-infrared bands (Irons et al., 2012), and improved signal-to-noise ratio and radiometric resolution make it superior for detecting sub-pixel spectral signatures of water (Pardo-Pascual et al., 2018). While previous studies have generally shown high agreement in per-band reflectance between the Landsat generations, the OLI sensor performs better on many downstream tasks (Mwaniki et al., 2015; Olmanson et al., 2016; Pardo-Pascual et al., 2018). Neural networks, which typically use nonlinear, continuous activation functions, may be especially sensitive to small differences. The effect of Landsat sensor differences on downstream tasks remains an important area for further research.

*4.3 Sources of Uncertainty*

Across all sensors, there are several potential sources of uncertainty. First, since most cloud-free imagery is obtained during the dry season (roughly May – September for most of the country), both the training data and final maps likely represent reservoir extent at or near its lowest point during the year. Thus, the reservoir sizes and counts derived from these maps are likely underestimates and do not fully capture variations in surface water throughout the years. In addition, using a random split can create spatial autocorrelation between the training, validation, and test sets, which can inflate model performance (Kattenborn et al., 2022). To test the effect of spatial autocorrelation between training, validation, and test data, we created Spatial Prediction Error Profiles (SPEPs; Brenning, 2023), assessing F1-score based on distance to the nearest training and validation tiles (Figure S1). While there is a slight drop in accuracy for test tiles over 100km from the nearest training tile (14 tiles), the overall results show that close proximity to training and validation tiles does not increase accuracy. In fact, accuracy for tiles 20 – 100 km from the nearest training set is higher than for tiles less than 20 km away. The overall effect of spatial autocorrelation appears minimal, perhaps due to the large area covered (8.5 million km$^2$) by 1510 annotated tiles—the median distance to the nearest training tile was 41 km.

In addition to confusion with natural water bodies, the model sometimes mistakes human-made rain-fed impoundments for in-stream reservoirs. In Brazil, Embrapa's barraginhas program has encouraged the construction of circular, rain-fed impoundments to mitigate increased runoff in agricultural landscapes (SEAMA - Espírito Santo, 2020), while avoiding some of the negative effects associated with in-stream reservoirs, including stream fragmentation and habitat disruption (Silva, 2021). The program was introduced in the 1990s in Minas Gerais, with wider adoption throughout the 2000s and 2010s (Landau et al., 2013; Rodrigo Maroni, 2019). If the model were falsely identifying large numbers of barraginhas as stream-fed reservoirs, we would expect to see an accelerating growth in reservoir counts during the 2000s and 2010s (particularly in Minas Gerais), but there is no such acceleration. Although the model misidentified some rain-fed impoundments as in-stream reservoirs, the distinctive circular shape of most barraginhas makes them relatively easy to distinguish. Nonetheless, users of our dataset should be aware of this potential source of confusion, particularly in Minas Gerais and other parts of the Caatinga and Cerrado where barraginhas are common.

## 5. Conclusions

Using four generations of Landsat sensors and deep learning computer vision, we produced consistent maps of small artificial water reservoirs (0.1 to 100 ha) in Brazil annually from 1984 to 2025, highlighting the potential of deep learning methods for mapping small-scale changes to global surface water over recent decades. Our results shed light on the dramatic expansion of stream barriers, a widespread environmental impact of agriculture in Brazil that has been understudied compared to deforestation. The publicly available annual reservoir maps can accelerate much-needed research to understand the specific drivers and impacts of reservoir construction and use.

## CRediT authorship contribution statement

**Kylen Solvik:** Conceptualization, Funding Acquisition, Methodology, Software, Writing–Original draft preparation. **Luis Carvalho:** Data Curation, Writing–Review & Editing. **Marcia Macedo:** Conceptualization, Funding Acquisition, Supervision, Resources, Writing–Review & Editing.

## Declaration of competing interests

The authors declare that they have no competing interests.

## Acknowledgments

We thank Margot McKlveen, Michael Lathuillière, Tomás Carvalho, Vivian Ribeiro, Nanxu Su, and Ylva Ran for their help annotating training data. This research was supported through grants from NASA (80NSSC23K1616; NNX16AI55G; 80NSSC23K0537) and NSF (MCS/DEB-1950832).

## Data availability statement

The complete reservoir dataset is available at: https://doi.org/10.5281/zenodo.21038980 (Solvik et al., 2026a). All code used for training, modeling, and analysis is available at: https://doi.org/10.5281/zenodo.21416344 (Solvik, 2026). The annotated training data and trained model weights are available at: https://doi.org/10.5281/zenodo.20450177 (Solvik et al., 2026b).

## References

ANA, 2024. Manual de Usos Consuntivos da Água no Brasil - 2ª Edição. Agência Nacional de Águas e Saneamento Básico.

ANA, 2020. Atualização da Base de Dados Nacional de Referência de Massas d'Água. Agência Nacional de Águas e Saneamento Básico.

Arvor, D., Daher, F.R.G., Briand, D., Dufour, S., Rollet, A.-J., Simões, M., Ferraz, R.P.D., 2018. Monitoring thirty years of small water reservoirs proliferation in the southern Brazilian Amazon with Landsat time series. ISPRS J. Photogramm. Remote Sens., SI: Latin America Issue 145, 225–237. https://doi.org/10.1016/j.isprsjprs.2018.03.015

Ballarin, A.S., Sousa Mota Uchôa, J.G., dos Santos, M.S., Almagro, A., Miranda, I.P., da Silva, P.G.C., da Silva, G.J., Gomes Júnior, M.N., Wendland, E., Oliveira, P.T.S., 2023. Brazilian Water Security Threatened by Climate Change and Human Behavior. Water Resour. Res. 59, e2023WR034914. https://doi.org/10.1029/2023WR034914

Barbarossa, V., Schmitt, R.J.P., Huijbregts, M.A.J., Zarfl, C., King, H., Schipper, A.M., 2020. Impacts of current and future large dams on the geographic range connectivity of freshwater fish worldwide. Proc. Natl. Acad. Sci. 117, 3648–3655. https://doi.org/10.1073/pnas.1912776117

Barona, E., Ramankutty, N., Hyman, G., Coomes, O.T., 2010. The role of pasture and soybean in deforestation of the Brazilian Amazon. Environ. Res. Lett. 5, 024002. https://doi.org/10.1088/1748-9326/5/2/024002

Bartuszevige, A., Pavlacky Jr, D., Burris, L., Herbener, K., 2012. Inundation of Playa Wetlands in the Western Great Plains Relative to Landcover Context. Wetlands 32. https://doi.org/10.1007/s13157-012-0340-6

Beaulieu, J.J., DelSontro, T., Downing, J.A., 2019. Eutrophication will increase methane emissions from lakes and impoundments during the 21st century. Nat. Commun. 10, 1375. https://doi.org/10.1038/s41467-019-09100-5

Belletti, B., Garcia de Leaniz, C., Jones, J., Bizzi, S., Börger, L., Segura, G., Castelletti, A., van de Bund, W., Aarestrup, K., Barry, J., Belka, K., Berkhuysen, A., Birnie-Gauvin, K., Bussettini, M., Carolli, M., Consuegra, S., Dopico, E., Feierfeil, T., Fernández, S., Fernandez Garrido, P., Garcia-Vazquez, E., Garrido, S., Giannico, G., Gough, P., Jepsen, N., Jones, P.E., Kemp, P., Kerr, J., King, J., Łapińska, M., Lázaro, G., Lucas, M.C., Marcello, L., Martin, P., McGinnity, P., O'Hanley, J., Olivo del Amo, R., Parasiewicz, P., Pusch, M., Rincon, G., Rodriguez, C., Royte, J., Schneider, C.T., Tummers, J.S., Vallesi, S., Vowles, A., Verspoor, E., Wanningen, H., Wantzen, K.M., Wildman, L., Zalewski, M., 2020. More than one million barriers fragment Europe's rivers. Nature 588, 436–441. https://doi.org/10.1038/s41586-020-3005-2

Brenning, A., 2023. Spatial machine-learning model diagnostics: a model-agnostic distance-based approach. Int. J. Geogr. Inf. Sci. 37, 584–606. https://doi.org/10.1080/13658816.2022.2131789

Brito, S.S.B., Cunha, A.P.M.A., Cunningham, C.C., Alvalá, R.C., Marengo, J.A., Carvalho, M.A., 2018. Frequency, duration and severity of drought in the Semiarid Northeast Brazil region. Int. J. Climatol. 38, 517–529. https://doi.org/10.1002/joc.5225

Cavalcante, L., Dewulf, A., Van Oel, P., 2022. Fighting against, and coping with, drought in Brazil: two policy paradigms intertwined. Reg. Environ. Change 22, 111. https://doi.org/10.1007/s10113-022-01966-4

Chandesris, A., Van Looy, K., Diamond, J.S., Souchon, Y., 2019. Small dams alter thermal regimes of downstream water. Hydrol. Earth Syst. Sci. 23, 4509–4525. https://doi.org/10.5194/hess-23-4509-2019

Chen, Y., Fan, R., Yang, X., Wang, J., Latif, A., 2018. Extraction of Urban Water Bodies from High-Resolution Remote-Sensing Imagery Using Deep Learning. Water 10, 585. https://doi.org/10.3390/w10050585

Crawford, C.J., Roy, D.P., Arab, S., Barnes, C., Vermote, E., Hulley, G., Gerace, A., Choate, M., Engebretson, C., Micijevic, E., Schmidt, G., Anderson, C., Anderson, M., Bouchard, M., Cook, B., Dittmeier, R., Howard, D., Jenkerson, C., Kim, M., Kleyians, T., Maiersperger, T., Mueller, C., Neigh, C., Owen, L., Page, B., Pahlevan, N., Rengarajan, R., Roger, J.-C., Sayler, K., Scaramuzza, P., Skakun, S., Yan, L., Zhang, H.K., Zhu, Z., Zahn, S., 2023. The 50-year Landsat collection 2 archive. Sci. Remote Sens. 8, 100103. https://doi.org/10.1016/j.srs.2023.100103

de Araújo, J.C., Döll, P., Güntner, A., Krol, M., Abreu, C.B.R., Hauschild, M., Mendiondo, E.M., 2004. Water Scarcity Under Scenarios for Global Climate Change and Regional Development in Semiarid Northeastern Brazil. Water Int. 29, 209–220. https://doi.org/10.1080/02508060408691770

DelSontro, T., Beaulieu, J.J., Downing, J.A., 2018. Greenhouse gas emissions from lakes and impoundments: Upscaling in the face of global change. Limnol. Oceanogr. Lett. 3, 64–75. https://doi.org/10.1002/lol2.10073

Dobrovolski, R., Rattis, L., 2015. Water collapse in Brazil: The danger of relying on what you neglect. Nat. Conserv. 345. https://doi.org/10.1016/j.ncon.2015.03.006

ESRI, Garmin, 2024. World Water Bodies [WWW Document]. URL https://hub.arcgis.com/content/e750071279bf450cbd510454a80f2e63 (accessed 3.25.24).

F. G. Assis, L.F., Ferreira, K.R., Vinhas, L., Maurano, L., Almeida, C., Carvalho, A., Rodrigues, J., Maciel, A., Camargo, C., 2019. TerraBrasilis: A Spatial Data Analytics Infrastructure for Large-Scale Thematic Mapping. ISPRS Int. J. Geo-Inf. 8, 513. https://doi.org/10.3390/ijgi8110513

Fan, T., Wang, G., Li, Y., Wang, H., 2020. MA-Net: A Multi-Scale Attention Network for Liver and Tumor Segmentation. IEEE Access 8, 179656–179665. https://doi.org/10.1109/ACCESS.2020.3025372

Fearnside, P., Figueiredo, A.M.R., Bonjour, S., 2013. Amazonian forest loss and the long reach of China’s influence. Environ. Dev. Sustain. 15. https://doi.org/10.1007/s10668-012-9412-2

Fearnside, P.M., 2005. Deforestation in Brazilian Amazonia: History, Rates, and Consequences. Conserv. Biol. 19, 680–688. https://doi.org/10.1111/j.1523-1739.2005.00697.x

Fearnside, P.M., 1982. Deforestation in the Brazilian Amazon: How fast is it occurring? Interciencia 7, 82–88.

Fencl, J.S., Mather, M.E., Costigan, K.H., Daniels, M.D., 2015. How Big of an Effect Do Small Dams Have? Using Geomorphological Footprints to Quantify Spatial Impact of Low-Head Dams and Identify Patterns of Across-Dam Variation. PLOS ONE 10, e0141210. https://doi.org/10.1371/journal.pone.0141210

Feng, W., Sui, H., Huang, W., Xu, C., An, K., 2019. Water Body Extraction From Very High-Resolution Remote Sensing Imagery Using Deep U-Net and a Superpixel-Based Conditional Random Field Model. IEEE Geosci. Remote Sens. Lett. 16, 618–622. https://doi.org/10.1109/LGRS.2018.2879492

Gharbia, R., 2023. Deep Learning for Automatic Extraction of Water Bodies Using Satellite Imagery. J. Indian Soc. Remote Sens. 51, 1511–1521. https://doi.org/10.1007/s12524-023-01705-0

Gorelick, N., Hancher, M., Dixon, M., Ilyushchenko, S., Thau, D., Moore, R., 2017. Google Earth Engine: Planetary-scale geospatial analysis for everyone. Remote Sens. Environ., Big Remotely Sensed Data: tools, applications and experiences 202, 18–27. https://doi.org/10.1016/j.rse.2017.06.031

Greenfield, G.M., 1992. The Great Drought and Elite Discourse in Imperial Brazil. Hisp. Am. Hist. Rev. 72, 375–400. https://doi.org/10.2307/2515990

Grinham, A., Albert, S., Deering, N., Dunbabin, M., Bastviken, D., Sherman, B., Lovelock, C.E., Evans, C.D., 2018. The importance of small artificial water bodies as sources of methane emissions in Queensland, Australia. Hydrol. Earth Syst. Sci. 22, 5281–5298. https://doi.org/10.5194/hess-22-5281-2018

Gui, S., Song, S., Qin, R., Tang, Y., 2024. Remote Sensing Object Detection in the Deep Learning Era—A Review. Remote Sens. 16, 327. https://doi.org/10.3390/rs16020327

Guo, M., Liu, H., Xu, Y., Huang, Y., 2020. Building Extraction Based on U-Net with an Attention Block and Multiple Losses. Remote Sens. 12, 1400. https://doi.org/10.3390/rs12091400

Habets, F., Philippe, E., Martin, E., David, C.H., Leseur, F., 2014. Small farm dams: impact on river flows and sustainability in a context of climate change. Hydrol. Earth Syst. Sci. 18, 4207–4222. https://doi.org/10.5194/hess-18-4207-2014

He, K., Zhang, X., Ren, S., Sun, J., 2016. Deep Residual Learning for Image Recognition, in: 2016 IEEE Conference on Computer Vision and Pattern Recognition (CVPR). Presented at the 2016 IEEE Conference on Computer Vision and Pattern Recognition (CVPR), IEEE, Las Vegas, NV, USA, pp. 770–778. https://doi.org/10.1109/CVPR.2016.90

Hirata, R., Goodarzi, L., Rörig, F.S., Alves, L.M., Bertolo, R., 2025. Climate change impacts on groundwater: a growing challenge for water resources sustainability in Brazil. Environ. Monit. Assess. 197, 784. https://doi.org/10.1007/s10661-025-14235-8

Holgerson, M.A., Raymond, P.A., 2016. Large contribution to inland water CO2 and CH4 emissions from very small ponds. Nat. Geosci. 9, 222–226. https://doi.org/10.1038/ngeo2654

Iakubovskii, P., 2019. Segmentation Models Pytorch. GitHub Repos.

Ioris, A.A.R., 2017. Encroachment and entrenchment of agro-neoliberalism in the Centre-West of Brazil. J. Rural Stud. 51, 15–27. https://doi.org/10.1016/j.jrurstud.2017.01.011

Irons, J.R., Dwyer, J.L., Barsi, J.A., 2012. The next Landsat satellite: The Landsat Data Continuity Mission. Remote Sens. Environ., Landsat Legacy Special Issue 122, 11–21. https://doi.org/10.1016/j.rse.2011.08.026

Jepson, W., Brannstrom, C., Filippi, A., 2010. Access Regimes and Regional Land Change in the Brazilian Cerrado, 1972–2002. Ann. Assoc. Am. Geogr. 100, 87–111. https://doi.org/10.1080/00045600903378960

Jonnala, N.S., Siraaj, S., Prastuti, Y., Chinnababu, P., Praveen babu, B., Bansal, S., Upadhyaya, P., Prakash, K., Faruque, M.R.I., Al-mugren, K.S., 2025. AER U-Net: attention-enhanced multi-scale residual U-Net structure for water body segmentation using Sentinel-2 satellite images. Sci. Rep. 15, 16099. https://doi.org/10.1038/s41598-025-99322-z

Kattenborn, T., Schiefer, F., Frey, J., Feilhauer, H., Mahecha, M.D., Dormann, C.F., 2022. Spatially autocorrelated training and validation samples inflate performance assessment of convolutional neural networks. ISPRS Open J. Photogramm. Remote Sens. 5, 100018. https://doi.org/10.1016/j.ophoto.2022.100018

Kibler, K.M., Tullos, D.D., 2013. Cumulative biophysical impact of small and large hydropower development in Nu River, China. Water Resour. Res. 49, 3104–3118. https://doi.org/https://doi.org/10.1002/wrcr.20243

Kim, J.H., Lee, H., Hong, S.J., Kim, S., Park, J., Hwang, J.Y., Choi, J.P., 2019. Objects Segmentation From High-Resolution Aerial Images Using U-Net With Pyramid Pooling Layers. IEEE Geosci. Remote Sens. Lett. 16, 115–119. https://doi.org/10.1109/LGRS.2018.2868880

Labelbox [WWW Document], 2025. URL https://labelbox.com (accessed 11.14.24).

Landau, E.C., Barros, L.C. de, Ribeiro, P.E. de A., Barros, I. de R., 2013. Abrangência geográfica do Projeto Barraginhas no Brasil. (No. 159), Embrapa Milho e Sorgo Documentos. Embrapa.

Lange, K., Meier, P., Trautwein, C., Schmid, M., Robinson, C.T., Weber, C., Brodersen, J., 2018. Basin-scale effects of small hydropower on biodiversity dynamics. Front. Ecol. Environ. 16, 397–404. https://doi.org/https://doi.org/10.1002/fee.1823

Lathuillière, M.J., Flach, R., Wang-Erlandsson, L., Ribeiro, V., zu Ermgassen, E.K.H.J., Souza, C.M., 2025. International reliance on Brazil’s water through soy and beef supply chains. Commun. Earth Environ. 6, 688. https://doi.org/10.1038/s43247-025-02658-7

Lathuillière, M.J., Solvik, K., Macedo, M.N., Graesser, J., Miranda, E.J., Couto, E.G., Johnson, M.S., 2019. Cattle production in Southern Amazonia: implications for land and water management. Environ. Res. Lett. 14, 114025. https://doi.org/10.1088/1748-9326/ab30a7

Luo, Y., Feng, A., Li, H., Li, D., Wu, X., Liao, J., Zhang, C., Zheng, X., Pu, H., 2022. New deep learning method for efficient extraction of small water from remote sensing images. PLOS ONE 17, e0272317. https://doi.org/10.1371/journal.pone.0272317

Macedo, M.N., Coe, M.T., DeFries, R., Uriarte, M., Brando, P.M., Neill, C., Walker, W.S., 2013. Land-use-driven stream warming in southeastern Amazonia. Philos. Trans. R. Soc. Lond. B. Biol. Sci. 368, 20120153. https://doi.org/10.1098/rstb.2012.0153

Mady, B., Lehmann, P., Gorelick, S.M., Or, D., 2020. Distribution of small seasonal reservoirs in semi-arid regions and associated evaporative losses. Environ. Res. Commun. 2, 061002.

https://doi.org/10.1088/2515-7620/ab92af

MapBiomas, 2025. Mapping of the Water Surface of Brazil - Collection 4.

Marengo, J.A., Torres, R.R., Alves, L.M., 2017. Drought in Northeast Brazil—past, present, and future. Theor. Appl. Climatol. 129, 1189–1200. https://doi.org/10.1007/s00704-016-1840-8

Morton, D.C., DeFries, R.S., Shimabukuro, Y.E., Anderson, L.O., Arai, E., del Bon Espirito-Santo, F., Freitas, R., Morisette, J., 2006. Cropland expansion changes deforestation dynamics in the southern Brazilian Amazon. Proc. Natl. Acad. Sci. 103, 14637–14641. https://doi.org/10.1073/pnas.0606377103

Mullen, A.L., Watts, J.D., Rogers, B.M., Carroll, M.L., Elder, C.D., Noomah, J., Williams, Z., Caraballo-Vega, J.A., Bredder, A., Rickenbaugh, E., Levenson, E., Cooley, S.W., Hung, J.K.Y., Fiske, G., Potter, S., Yang, Y., Miller, C.E., Natali, S.M., Douglas, T.A., Kyzivat, E.D., 2023. Using High-Resolution Satellite Imagery and Deep Learning to Track Dynamic Seasonality in Small Water Bodies. Geophys. Res. Lett. 50, e2022GL102327. https://doi.org/10.1029/2022GL102327

Nóbrega, M.T., Collischonn, W., Tucci, C.E.M., Paz, A.R., 2011. Uncertainty in climate change impacts on water resources in the Rio Grande Basin, Brazil. Hydrol. Earth Syst. Sci. 15, 585–595. https://doi.org/10.5194/hess-15-585-2011

Odebiri, O., Archbold, J., Glen, J., Macreadie, P.I., Malerba, M.E., 2024. Excluding livestock access to farm dams reduces methane emissions and boosts water quality. Sci. Total Environ. 951, 175420. https://doi.org/10.1016/j.scitotenv.2024.175420

Oliveira, G., Hecht, S., 2016. Sacred groves, sacrifice zones and soy production: globalization, intensification and neo-nature in South America. J. Peasant Stud. 43, 251–285. https://doi.org/10.1080/03066150.2016.1146705

Ollivier, Q.R., Maher, D.T., Pitfield, C., Macreadie, P.I., 2019. Punching above their weight: Large release of greenhouse gases from small agricultural dams. Glob. Change Biol. 25, 721–732. https://doi.org/10.1111/gcb.14477

Pardo-Pascual, J.E., Sánchez-García, E., Almonacid-Caballer, J., Palomar-Vázquez, J.M., Priego de los Santos, E., Fernández-Sarría, A., Balaguer-Beser, Á., 2018. Assessing the Accuracy of Automatically Extracted Shorelines on Microtidal Beaches from Landsat 7, Landsat 8 and Sentinel-2 Imagery. Remote Sens. 10, 326. https://doi.org/10.3390/rs10020326

Peacock, M., Audet, J., Bastviken, D., Cook, S., Evans, C.D., Grinham, A., Holgerson, M.A., Högbom, L., Pickard, A.E., Zieliński, P., Futter, M.N., 2021. Small artificial waterbodies are widespread and persistent emitters of methane and carbon dioxide. Glob. Change Biol. 27, 5109–5123. https://doi.org/10.1111/gcb.15762

Pekel, J.-F., Cottam, A., Gorelick, N., Belward, A.S., 2016. High-resolution mapping of global surface water and its long-term changes. Nature 540, 418–422. https://doi.org/10.1038/nature20584

Pendrill, F., Gardner, T.A., Meyfroidt, P., Persson, U.M., Adams, J., Azevedo, T., Bastos Lima, M.G., Baumann, M., Curtis, P.G., De Sy, V., Garrett, R., Godar, J., Goldman, E.D., Hansen, M.C., Heilmayr, R., Herold, M., Kuemmerle, T., Lathuillière, M.J., Ribeiro, V., Tyukavina, A., Weisse, M.J., West, C., 2022. Disentangling the numbers behind agriculture-driven tropical deforestation. Science 377, eabm9267. https://doi.org/10.1126/science.abm9267

Perazzoli, M., Pinheiro, A., Kaufmann, V., 2013. Assessing the impact of climate change scenarios on water resources in southern Brazil. Hydrol. Sci. J. 58, 77–87. https://doi.org/10.1080/02626667.2012.742195

Pi, X., Luo, Q., Feng, L., Xu, Y., Tang, J., Liang, X., Ma, E., Cheng, R., Fensholt, R., Brandt, M., Cai, X., Gibson, L., Liu, J., Zheng, C., Li, W., Bryan, B.A., 2022. Mapping global lake dynamics reveals the emerging roles of small lakes. Nat. Commun. 13, 5777. https://doi.org/10.1038/s41467-022-33239-3

Rattis, L., Brando, P.M., Macedo, M.N., Spera, S.A., Castanho, A.D.A., Marques, E.Q., Costa, N.Q., Silverio, D.V., Coe, M.T., 2021. Climatic limit for agriculture in Brazil. Nat. Clim. Change 1–7. https://doi.org/10.1038/s41558-021-01214-3

Rausch, L.L., Gibbs, H.K., 2016. Property Arrangements and Soy Governance in the Brazilian State of Mato Grosso: Implications for Deforestation-Free Production. Land 5, 7. https://doi.org/10.3390/land5020007

Rodrigo Maroni, J., 2019. Ele “plantou” chuva no sertão e melhorou a vida de milhares de pessoas. Gaz. Povo.

Schulz, C., Ioris, A.A.R., 2017. The Paradox of Water Abundance in Mato Grosso, Brazil. Sustainability 9, 1796. https://doi.org/10.3390/su9101796

SEAMA - Espírito Santo, 2020. Projeto Barraginhas está ajudando a combater a desertificação em municípios capixabas [WWW Document]. Secr. Meio Ambiente E Recur. Hídricos SEAMA - Espírito St. URL https://seama.es.gov.br/Notícia/projeto-barraginhas-esta-ajudando-a-combater-a-desertificacao-em-municipios-capixabas (accessed 1.19.25).

Shen, H., Mellempudi, N., He, X., Gao, Q., Wang, C., Wang, M., 2023. Efficient Post-training Quantization with FP8 Formats. https://doi.org/10.48550/arXiv.2309.14592

Silva, M.C.A. da, 2021. Impactos das barraginhas: uma tecnologia social no cotidiano de famílias do assentamento Rancho da Saudade, no município de Cáceres-MT. Editora Unemat, Cáceres, MT.

Silva Junior, C.H.L., Pessôa, A.C.M., Carvalho, N.S., Reis, J.B.C., Anderson, L.O., Aragão, L.E.O.C., 2021. The Brazilian Amazon deforestation rate in 2020 is the greatest of the decade. Nat. Ecol. Evol. 5, 144–145. https://doi.org/10.1038/s41559-020-01368-x

Solvik, K., 2026. kysolvik/reservoir-id-smp: v0.3.2. https://doi.org/10.5281/zenodo.21416344

Solvik, K., Bartuszevige, A.M., Bogaerts, M., Joseph, M.B., 2021. Predicting Playa Inundation Using a Long Short-Term Memory Neural Network. Water Resour. Res. 57, e2020WR029009. https://doi.org/10.1029/2020WR029009

Solvik, K., Carvalho, L.G., Nunes Macedo, M., 2026a. Reservoir maps from “Mapping small reservoirs across Brazil from 1984 to 2025.” https://doi.org/10.5281/zenodo.21038980

Solvik, K., Carvalho, L.G., Nunes Macedo, M., 2026b. Model weights and training, validation, and test set images and masks for “Mapping the evolution of small reservoirs in Brazil from 1984-2025 using deep learning.” https://doi.org/10.5281/zenodo.20450177

Souza, C.M., Kirchhoff, F.T., Oliveira, B.C., Ribeiro, J.G., Sales, M.H., 2019. Long-Term Annual Surface Water Change in the Brazilian Amazon Biome: Potential Links with Deforestation, Infrastructure Development and Climate Change. Water 11, 566. https://doi.org/10.3390/w11030566

Sun, D., Gao, G., Huang, L., Liu, Y., Liu, D., 2024. Extraction of water bodies from high-resolution remote sensing imagery based on a deep semantic segmentation network. Sci. Rep. 14, 14604.

https://doi.org/10.1038/s41598-024-65430-5

Sun, J., Ding, C., Lucas, M.C., Tao, J., Cheng, H., Chen, J., Li, M., Ding, L., Ji, X., Wang, Y., He, D., 2024. Convolutional Neural Networks Facilitate River Barrier Detection and Evidence Severe Habitat Fragmentation in the Mekong River Biodiversity Hotspot. Water Resour. Res. 60, e2022WR034375. https://doi.org/10.1029/2022WR034375

Sun, J., Lucas, M.C., Olden, J.D., Couto, T.B.A., Ning, N., Duffy, D., Baumgartner, L.J., 2025. Towards a comprehensive river barrier mapping solution to support environmental management. Nat. Water 3, 38–48. https://doi.org/10.1038/s44221-024-00364-w

Tanny, J., Cohen, S., Assouline, S., Lange, F., Grava, A., Berger, D., Teltch, B., Parlange, M.B., 2008. Evaporation from a small water reservoir: Direct measurements and estimates. J. Hydrol. 351, 218–229. https://doi.org/10.1016/j.jhydrol.2007.12.012

Trevisiol, F., Mandanici, E., Pagliarani, A., Bitelli, G., 2024. Evaluation of Landsat-9 interoperability with Sentinel-2 and Landsat-8 over Europe and local comparison with field surveys. ISPRS J. Photogramm. Remote Sens. 210, 55–68. https://doi.org/10.1016/j.isprsjprs.2024.02.021

Wieland, M., Martinis, S., Kiefl, R., Gstaiger, V., 2023. Semantic segmentation of water bodies in very high-resolution satellite and aerial images. Remote Sens. Environ. 287, 113452. https://doi.org/10.1016/j.rse.2023.113452

Xu, H., Ren, M., Lin, M., 2024. Cross-comparison of Landsat-8 and Landsat-9 data: a three-level approach based on underfly images. GIScience Remote Sens. 61, 2318071. https://doi.org/10.1080/15481603.2024.2318071

Yuan, K., Zhuang, X., Schaefer, G., Feng, J., Guan, L., Fang, H., 2021. Deep-Learning-Based Multispectral Satellite Image Segmentation for Water Body Detection. IEEE J. Sel. Top. Appl. Earth Obs. Remote Sens. 14, 7422–7434. https://doi.org/10.1109/JSTARS.2021.3098678

Zaidel, P.A., Roy, A.H., Houle, K.M., Lambert, B., Letcher, B.H., Nislow, K.H., Smith, C., 2021. Impacts of small dams on stream temperature. Ecol. Indic. 120, 106878. https://doi.org/10.1016/j.ecolind.2020.106878

Zhang, W., Pan, H., Song, C., Ke, L., Wang, J., Ma, R., Deng, X., Liu, K., Zhu, J., Wu, Q., 2019. Identifying Emerging Reservoirs along Regulated Rivers Using Multi-Source Remote Sensing

Observations. Remote Sens. 11, 25. https://doi.org/10.3390/rs11010025

Zhang, Z., Xia, X., Liu, S., Wang, J., Xu, W., McDowell, W.H., Tian, H., Yang, Z., 2026. Human-impacted lakes contribute over one-third of global lake greenhouse gas emissions. One Earth 9, 101568. https://doi.org/10.1016/j.oneear.2025.101568

## Supplementary Material

### S1. Mask annotation and data preprocessing

To create Landsat and Sentinel-2 composites, we used a modified version of Earth Engine's Landsat "simpleComposite" method, which selects low-cloud pixels and calculates median composites (Pekel et al., 2016). While the original implementation used top-of-atmosphere reflectance, we used Landsat Collection 2, Level 2 Tier 1 surface reflectance data, which offers improved consistency between sensors (Crawford et al., 2023). For each year, we included all Landsat scenes from the target year and adjacent years to minimize cloud contamination. For example, the inputs for the 1990 composite included scenes from 1989-1991.

To create annotated reservoir masks for training and evaluation, we manually outlined every reservoir in randomly selected 5km x 5km subsets using LabelBox based on Sentinel-2 NDWI and high-resolution imagery (Google Earth satellite and airborne imagery). Two criteria were used for distinguishing on-stream reservoirs from other waterbodies during annotation: (1) Is there a visible dam or other obstruction (e.g. road) impounding the water? This is typically evidenced by an unnaturally straight line on the downstream edge of the waterbody. (2) Is there evidence of a stream path flowing into and/or out of the reservoir? This is usually indicated by a riverine path with higher NDWI than the surroundings, even when the stream is dry. During annotation, subset images were presented for labeling in random order. However, in cases where the human annotator could not accurately and confidently identify the reservoirs, the image was skipped to avoid introducing bad training data to the model, which may create a positive bias in our accuracy metrics. In total, 1314 images were labeled in this way, containing a total of 4474 reservoirs. With the 196 extra floodplains and wetlands tiles, we had 1510 masks from which to create training, validation, and test sets. Note that because the project first began with a focus on the Amazon/Cerrado transition zone, this area is overrepresented in the annotated data—53% of the annotated images come from this region, which covers 32% of Brazil's land area.

Before training, to minimize the effects of high-value outliers, we calculated the maximum values for each band over the 1510 tiles and capped values at the 80th percentile. We then normalized all pixel values to a sample mean of 0 and standard deviation of 1.

**S2. Inference, post-processing, and analysis**

We ran model inference over all of Brazil by dividing the full composites into 360,504 tiles, each spanning 640 x 640 pixels with a stride of 480 pixels, to produce final clipped outputs that overlapped by 10 pixels, reducing the effect of missing context at image edges. We ran inference for Landsat 5 from 1984 - 2011, Landsat 7 from 2000 - 2024, Landsat 8 from 2014 – 2025, and Landsat 9 from 2022 – 2025. Note that we excluded the first year of Landsat 7 and 8 (1999 and 2013, respectively) due to insufficient data availability. Lightweight Google Compute Engine instances with two AMD EPYC Milan virtual CPUs performed the inference. To accelerate the computations, we applied integer quantization to the model using Intel Neural Compressor (Shen et al., 2023), which resulted in a negligible decrease in validation accuracy but 50% higher throughput. After inference, we mosaicked the tiles into full, wall-to-wall maps for each year and satellite combination, resulting in a total of 69 binary reservoir maps over the 42-year time period.

After the reservoir maps were created, we removed cloud-corrupted pixels and false positives in known large waterbodies. First, we assigned each pixel the maximum cloud score found within a 100 m radius of it, since the neural network model depends heavily on local context to make pixel-wise predictions. Then, we masked out pixels with cloud scores greater than 70 (out of 100). This removed less than 0.3% of pixels across all years and satellites. In some cases, the model falsely identified segments of very large waterbodies (e.g., inlets or bays on lakes or large hydroelectric reservoirs) as small reservoirs. Since this was rare and only occurred for large, permanent waterbodies, we filtered out reservoir detections that overlapped with a known water body larger than 100 ha from the ESRI/Garmin World Water Bodies database (ESRI and Garmin, 2024). This removed 0.4% of reservoir detections. To provide an unbiased representation of the model's performance, test-set accuracy metrics were evaluated before this step.

We evaluated the accuracy of the final model against the withheld test set of 263 annotated images. We calculated accuracy statistics on both a per-pixel (i.e. each pixel is evaluated as a true positive, true negative, false positive, or false negative) and per-object basis (i.e. each reservoir *object* is evaluated). For the object-based scores, a True Positive was considered any predicted reservoir object that intersected with a ground-truth object. In cases where multiple predictions overlapped with a single ground-truth object, only the largest prediction was considered a True Positive and all others were classified as False Positives. We calculated object-based precision and recall for four different size classes (below 0.1 ha, 0.1 to 1 ha, 1 to 10 ha, 10 to 100 ha; based on the size of the prediction mask at 10m resolution). Based on these results, we removed reservoir detections smaller than 0.1 ha due to low accuracy (precision = 47.2% and recall = 6.9%) from the final maps.

**Table S1.** Accuracy statistics on 2017 validation set, comparing sensors and resampling methods.

| | **Landsat 8 (10m, bicubic)** | **Landsat 8 (10m, bilinear)** | **Landsat 8 (native 30m)** | **Landsat 7 (10m, bicubic)** |
|---|---|---|---|---|
| **IoU-Score** | 54.8 | 53.7 | 49.6 | 54.3 |
| **F1-Score** | 70.8 | 69.9 | 66.3 | 70.3 |
| **Precision** | 70.7 | 69.5 | 66.5 | 69.8 |
| **Recall** | 70.8 | 70.2 | 66.0 | 70.8 |

**Table S2.**
**Accuracy by reservoir size (object-based). Highest values for each category are shown in bold.**

| | | | *Landsat 7* | | | *Landsat 8* | | |
|---|---|---|---|---|---|---|---|---|
| *Split* | *Size (ha)* | *True Count* | *Predicted Count* | *Precision* | *Recall* | *Predicted Count* | *Precision* | *Recall* |
| *Val* | < 0.1* | 337 | 253 | 4.4 | 3.3 | 96 | 28.1 | 8.0 |
| | 0.1 to 1 | 486 | 226 | 76.1 | 35.4 | 369 | 69.1 | 52.5 |
| | 1 to 10 | 76 | 87 | 80.5 | 92.1 | 81 | 86.4 | 92.1 |
| | 10 to 100 | 13 | 15 | 80.0 | 92.3 | 13 | 92.3 | 92.3 |
| | All | 575 | 328 | 80.8 | 46.1 | 463 | 73.0 | 58.8 |
| *Test* | < 0.1* | 422 | 256 | 3.9 | 2.4 | 89 | 32.6 | 6.9 |
| | 0.1 to 1 | 488 | 244 | 81.1 | 40.6 | 379 | 76.0 | 59.0 |
| | 1 to 10 | 85 | 107 | 69.2 | 87.1 | 105 | 73.3 | 90.6 |
| | 10 to 100 | 7 | 10 | 50.0 | 71.4 | 10 | 50.0 | 71.4 |
| | All | 580 | 361 | 83.7 | 52.1 | 494 | 78.7 | 67.1 |

*****Removed from "All" and from final maps.

**Table S3. Total reservoir surface area and count, 1984 - 2025**

| Year | Count | Surface Area ($km^2$) |
|---|---|---|
| 1984 | 224,059 | 348,987 |
| 1985 | 244,118 | 386,113 |
| 1986 | 251,274 | 388,538 |
| 1987 | 261,572 | 398,955 |
| 1988 | 270,053 | 407,276 |
| 1989 | 294,469 | 438,106 |
| 1990 | 277,484 | 402,088 |
| 1991 | 246,074 | 357,039 |
| 1992 | 218,981 | 318,240 |
| 1993 | 242,089 | 346,248 |
| 1994 | 267,996 | 371,283 |
| 1995 | 316,079 | 444,739 |
| 1996 | 326,884 | 456,351 |
| 1997 | 305,552 | 413,773 |
| 1998 | 313,148 | 412,831 |
| 1999 | 336,174 | 435,826 |
| 2000 | 413,804 | 532,149 |
| 2001 | 439,332 | 558,127 |
| 2002 | 433,800 | 555,622 |
| 2003 | 457,636 | 591,622 |
| 2004 | 467,229 | 606,528 |
| 2005 | 489,612 | 628,049 |
| 2006 | 496,087 | 631,604 |
| 2007 | 512,424 | 656,759 |
| 2008 | 543,209 | 707,783 |
| 2009 | 551,795 | 720,387 |
| 2010 | 557,965 | 726,005 |
| 2011 | 522,159 | 676,318 |
| 2012 | 518,188 | 673,671 |
| 2013 | 521,738 | 658,843 |
| 2014 | 614,141 | 673,956 |
| 2015 | 611,831 | 669,826 |
| 2016 | 614,953 | 674,947 |
| 2017 | 655,201 | 714,580 |
| 2018 | 679,642 | 753,621 |
| 2019 | 692,107 | 774,344 |
| 2020 | 797,728 | 801,561 |
| 2021 | 821,656 | 813,205 |
| 2022 | 826,060 | 820,268 |
| 2023 | 832,325 | 848,483 |
| 2024 | 830,725 | 860,992 |
| 2025 | 817,458 | 845,545 |

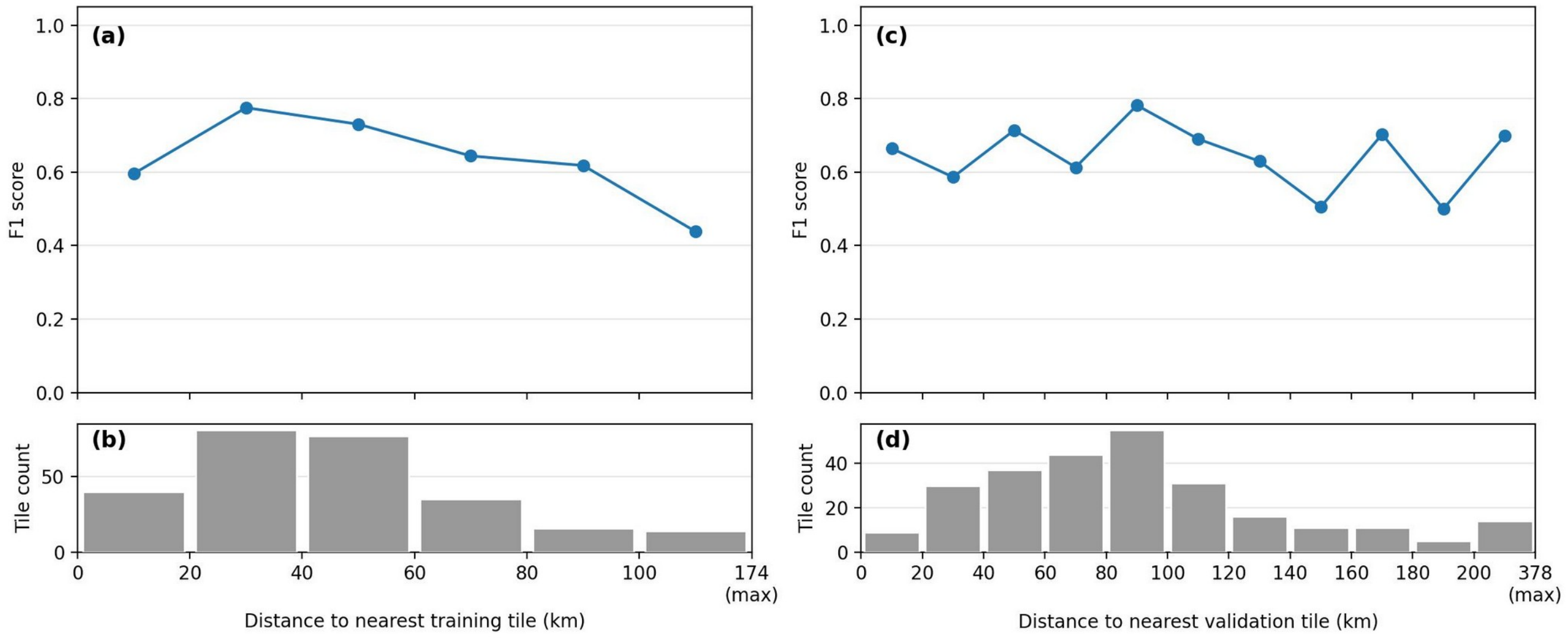

**Figure S1. Test set F1-score grouped by distance from nearest training and validation tiles.** Spatial autocorrelation between training, validation, and test set locations does not appear to be a major influence on test set accuracy.